\documentclass[journal]{IEEEtran}

\usepackage[utf8]{inputenc}

\usepackage{graphicx}
\usepackage{amsmath}
\usepackage[version=4]{mhchem}
\usepackage{longtable,tabularx}
\usepackage{tikz}
\usepackage[square,sort,comma,numbers]{natbib}
\usepackage{hyperref}
\usepackage{amssymb}
\usepackage{booktabs} 
\usepackage{siunitx}    
\usepackage{interval}   

\DeclareMathOperator*{\argmin}{arg\,min} 

\makeatletter
\newcommand\mathcircled[1]{%
  \mathpalette\@mathcircled{#1}%
}
\newcommand\@mathcircled[2]{%
  \tikz[baseline=(math.base)] \node[draw,circle,inner sep=1pt] (math) {$\m@th#1#2$};%
}
\makeatother

\usepackage{changes}
\definechangesauthor[name={Dharmesh Tailor}, color=red]{DT}
\definechangesauthor[color=black!50!green, name={Language/Grammar}]{lang}

\newcommand{\finalstate}[1][]{%
  \renewcommand{\added}[2][]{##2}%
  \renewcommand{\replaced}[3][]{##2}}%

\usepackage[draft]{todonotes}   

\begin{document}
%
\title{On the stability analysis of deep neural network representations of an optimal state-feedback}

%
%
%

\author{Dario~Izzo,
        Dharmesh~Tailor,
        and~Thomas~Vasileiou
\thanks{All authors are with the Advanced Concepts Team of the European Space Agency. Keplerlaan, 2201AZ, Noordwijk, The Netherlands.}}

\maketitle


\begin{abstract}
Recent work have shown how the optimal state-feedback, obtained as the solution to the Hamilton-Jacobi-Bellman equations, can be approximated for several nonlinear, deterministic systems by deep neural networks. 
When imitation (supervised) learning is used to train the neural network on optimal state-action pairs, for instance as derived by applying Pontryagin's theory of optimal processes, the resulting model is referred here as the guidance and control network. 
In this work, we analyze the stability of nonlinear and deterministic systems controlled by such networks. 
We then propose a method utilising differential algebraic techniques and high-order Taylor maps to gain information on the stability of the neurocontrolled state trajectories. 
We exemplify the proposed methods in the case of the two-dimensional dynamics of a quadcopter controlled to reach the origin and we study how different architectures of the guidance and control network affect the stability of the target equilibrium point and the stability margins to time delay.
Moreover, we show how to study the robustness to initial conditions of a nominal trajectory, using a Taylor representation of the neurocontrolled neighbouring trajectories.
\end{abstract}


%
\IEEEpeerreviewmaketitle

\section{Introduction}

The optimal feedback of several deterministic, non-linear systems of interest in aerospace applications has been, recently, directly represented by deep neural networks trained using techniques such as imitation learning \citep{sanchez2018real, tailor2019learning, mcdermott1994approximating, furfaro2018recurrent} or reinforcement learning \citep{levine2013exploring, zhang15, vamvoudakis2010online}. Regardless of the training details, the neural network is approximating the solution to the Hamilton-Jacobi-Bellman (HJB) equations.
The revival of interest in such methods is due to recent advances in deep learning, not limited to: learning algorithms, regularisation techniques, exploitation of GPUs for faster training and large datasets.
%

The term G\&CNET (Guidance and Control Network) is here used to refer to one such representation, in particular to a feedforward, fully-connected neural network trained using supervised learning to approximate the optimal feedback---a function relating the state to the optimal action---of an autonomous, deterministic system. 
The optimal state-action pairs that constitute the training set are computed by applying Pontryagin's theory of optimal processes \citep{pontryagin1987mathematical}, or by solving the HJB equations. 
In other words, G\&CNETs are neural networks imitating the optimal feedback of a nonlinear, autonomous and deterministic system, where the word \lq\lq imitating\rq\rq\ has to be taken in the sense of the machine learning technique of imitation learning \citep{ho2016generative} of expert's actions.
G\&CNETs can be viewed as an alternative to the widely used two-degrees-of-freedom approach to optimal control, based on tracking a pre-computed optimal guidance profile. In the two-degrees-of-freedom approach, the guidance and control problem is decoupled into two tasks: trajectory generation, which is done offline, and the trajectory tracking, which is taken care by an on-board controller or a deep neural network.
On the other hand, in the case of G\&CNETs both tasks are performed by the neural network in real-time; the remainder of the trajectory is uniquely defined from the current state and perturbations from a nominal trajectory result in alternative profiles.
The idea behind G\&CNETs is not new and similar schemes were studied in the 90s \citep{mcdermott1994approximating}. 
It is, though, only in recent years, thanks to advances in the use of neural networks and computer hardware, that the idea was successfully deployed on higher dimensional and more complex systems and thus received renewed attention \citep{furfaro2018recurrent, sanchez2018real, tailor2019learning}.

Neurocontrollers have been studied in the past and results on their stability obtained in several cases \cite{vamvoudakis2010online, vamvoudakis2014neural, nodland2013neural, hrycej1995stability}.
In particular, stable neurocontrollers can be designed by applying techniques from the adaptive control framework \cite{Aastrom2013_ch5}; the adaption of the neurocontroller weights is performed in real-time satisfying a predefined Lyapunov function \cite{vamvoudakis2014neural, nodland2013neural}.
Although training of the neural network is entirely avoided, the technique is only applicable to shallow neural networks.
Separately, the notion of stability around a nominal trajectory used to analyze controllers which are developed to perform the tracking task in a two-degrees-of-freedom design, cannot be transferred to G\&CNETs.
Perturbations along a nominal trajectory are not driven to zero, but drift as the neurocontroller applies the approximate learned optimal feedback.
Furthermore, the generally perceived black-box nature of neural networks leads to skepticism on the use of such controllers.
Especially in cases where safety and validation are of paramount importance, such as the automotive or space industry, other solutions are employed even if neural networks would provide competitive performance. 

In this paper we first analyze the linear stability of a system controlled by a G\&CNET and then we propose a new method, based on the use of differential algebra and high-order Taylor maps (HOTM) \citep{berz1998verified}, to gain information on the stability of neurocontrolled state trajectories with respect to their initial conditions.
After presenting a general methodology, we introduce the case study of a two-dimensional quadcopter dynamics. 
We train several G\&CNETs with different architectures to approximate the power-optimal control of the system (to reach a hovering equilibrium at the origin) and we study their linear behaviour in proximity to the equilibrium point.
We then show how to use Taylor models to describe the system behaviour around any neurocontrolled trajectory and obtain stability estimates

\footnote{The code and data to reproduce the results in this paper is made available via the github repository \url{https://github.com/darioizzo/neurostability}.}

\section{Methodology}
\label{sec:methods}
In this section we state formally the optimal control problem considered and introduce the formalism used to construct neural network representations of the optimal state-feedback or G\&CNETs

Later we present the linear stability study of a neurocontrolled system highlighting the explicit relation between the linearized system and neural network gradients. 
We then introduce the use of a high-order Taylor model of a neurocontrolled trajectory in proximity of a reference trajectory to study its stability with respect to perturbations of the initial conditions. 

\subsection{Optimal Control problem statement}
We study the non-linear, autonomous system $\dot{\mathbf x} = \mathbf f(\mathbf x, \mathbf u)$, where $\mathbf x(\cdot): \mathbb{R}\rightarrow \mathbb{R}^n$, $\mathbf u(\cdot): \mathbb{R} \rightarrow \mathbb{R}^m$ and $\mathbf f(\cdot,\cdot): \mathbb{R}^n \times \mathbb{R}^m \rightarrow \mathbb{R}^n$. 
The free terminal time optimal control problem is then introduced as the problem of finding a control $\mathbf u(\cdot) \in \mathcal U$ able to steer in the time $t_f$ such a system from an initial state $\mathbf x_0$ to a target state $\mathbf x_f$ minimizing the cost functional $J(\mathbf x, \mathbf u, t_f) = \int_{t_0}^{t_f}\ell(\mathbf x, \mathbf u)dt$. 
The functional space $\mathcal U$ is the space of all piecewise continuous functions assuming values in some closed region $U \subset \mathbb{R}^m$. 
The relationship between the optimal value attained by $J$ and the initial state $\mathbf x_0$ is called the value function and is here indicated with $v(\mathbf x_0)$.
We restrict our attention to problems in which the cost rate $\ell$ does not depend on time. This means the HJB equations have a time-invariant form.
In all points $\mathbf x$ where the value function is differentiable, the optimal control is given by \cite{todorov2006optimal}:
\begin{equation}\label{eq:hjb_policy}
\mathbf u^*(\mathbf x) = \argmin_{\mathbf u \in \mathcal U}\left\{ \ell(\mathbf x, \mathbf u) + \mathbf f(\mathbf x, \mathbf u) \boldsymbol{\cdot} \nabla_{\mathbf{x}}v(\mathbf x)\right\}.
\end{equation}
%
Beyond classic solutions, the optimal value function can be given uniquely by considering the viscosity solution to the HJB equations \cite{crandall1992user}. Furthermore, if equation (\ref{eq:hjb_policy}) has a single minimizer, the optimal control problem also admits a unique solution $\mathbf u^*(\mathbf x)$ called the optimal feedback. 
The relation above reveals how the optimal control for such systems is purely reactive, as it depends only on the current system state and not on its history. 
When controlled by the optimal feedback the system will acquire the desired target state at $t^*_f$. The resulting optimal trajectories are denoted by $\mathbf x^*(t)$. 

\subsection{Neural network representation of the optimal feedback}
Fundamental problems in aerospace engineering, such as low-thrust spacecraft interplanetary transfers, spacecraft landing, unmanned aerial vehicle control and rocket guidance, all fall into the description above. Finding a solution to these problems is known to require significant computational resources. 
This is due to the complex structure of the resulting control problem as well as the mathematical and numerical issues connected with the HJB equation or with the application of Pontryagin's maximum principle.
As a consequence, approximating $\mathbf u^*(\mathbf x)$, or the value function $v(\mathbf x)$, is desirable and proposals have been widely researched in the past both non-neural network based \cite{beard1997galerkin} and neural network based \cite{vamvoudakis2014neural, levine2013exploring, levine13guidedpolicy, sanchez2018real}. 
We take the approach discussed in \cite{sanchez2018real} and study neural networks, with feedforward, fully-connected architectures, trained on a database of optimal state-action pairs computed by solving, from different initial conditions $\mathbf x_0$, the two-point boundary value problem arising from the application of Pontryagin's maximum principle \citep{pontryagin1987mathematical}. 
The resulting trained networks are here indicated with the term G\&CNET as they can be used on-board in real time as a substitute for more classical guidance and control systems.

Even before training commences, we can infer several properties of the G\&CNET by looking at the state-control pairs comprising the database.
One such property is their behaviour after reaching the target state $\mathbf x_f$; critical for the use of these controllers on-board.
This is strongly influenced by the controls associated with $\mathbf x_f$, which is by far the majority state in the database as it appears once for each optimal trajectory. 
In particular, the value function gradient $\nabla_{\mathbf{x}}v(\mathbf x)$ is not uniquely defined at the target point $\mathbf x_f$ and neither is the corresponding optimal control $\mathbf u^*(\mathbf x_f)$. As a consequence, depending on the initial conditions chosen, the training database will contain contradictory state-action pairs defining the policy in $\mathbf x_f$. In most cases the neural network will learn to associate a value close to the average control at $\mathbf x_f$.
Whilst manipulation of the initial conditions used in trajectory generation is a viable approach as done in \citep{tailor2019learning}, we propose to overwrite $\mathbf u^*(\mathbf x_f)$ as a post-processing step following database creation.
This is in the spirit of \citep{mcdermott1994approximating} which proposed the exploitation of \textit{a priori} knowledge (i.e. the specific control problem) when designing the controller.
Proposing a suitable $\mathbf u^*(\mathbf x_f)$ is made easier when we restrict our attention to optimal control problems where $\mathbf x_f$ is also an equilibrium point for the system under some control $\mathbf u_e$, that is we assume $\exists ! \mathbf u_e$ such that $\mathbf f(\mathbf x_f, \mathbf u_e) = \mathbf 0$. 
Consequently, we extend the optimal feedback definition by setting $\mathbf u^*(\mathbf x_f) = \mathbf u_e$. From this point onwards we will use $\mathbf x_e$ to denote the final target state. Note that $\mathbf x_e$ is a globally asymptotically stable equilibrium point for the optimally controlled system $\dot{\mathbf x} = \mathbf f(\mathbf x, \mathbf u^*)$, as can be promptly shown realizing that the value function $v(\mathbf x)$ exists unique and is a Lyapunov function for the system (every meaningful value function is a Lyapunov function and every Lyapunov function is a meaningful value function \cite{freeman1996inverse}).

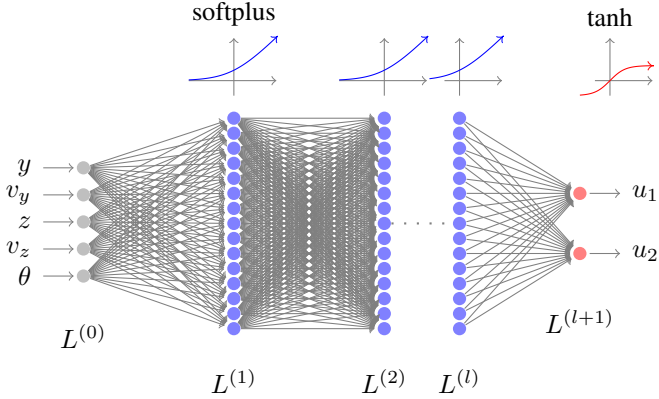
\begin{figure}[tb]
\centering
\def\layersep{2cm}
\def\nodesep{0.2cm}

\begin{tikzpicture}[shorten >=1pt,->,draw=black!50, node distance=\layersep]
    \tikzstyle{every pin edge}=[<-,shorten <=1pt]
    \tikzstyle{neuron}=[circle,fill=black!25,minimum size=5pt,inner sep=0pt]
    \tikzstyle{input neuron}=[neuron, fill=gray!50];
    \tikzstyle{output neuron}=[neuron, fill=red!50];
    \tikzstyle{hidden neuron softplus}=[neuron, fill=blue!50];

    \tikzstyle{annot} = [text width=4em, text centered]

    \node[input neuron, pin=left:$y$] (I-1) at (0,-\nodesep*1*1.8) {};
    \node[input neuron, pin=left:$v_y$] (I-2) at (0,-\nodesep*2*1.8) {};
    \node[input neuron, pin=left:$z$] (I-3) at (0,-\nodesep*3*1.8) {};
    \node[input neuron, pin=left:$v_z$] (I-4) at (0,-\nodesep*4*1.8) {};
    \node[input neuron, pin=left:$\theta$] (I-5) at (0,-\nodesep*5*1.8) {};
    \node[] (L0) at (0,-\nodesep*5*1.8-4*\nodesep) {$L^{(0)}$};

    \foreach \name / \y in {1,...,15}
        \path[yshift=0.5cm]
            node[hidden neuron softplus] (H1-\name) at (\layersep,-\nodesep*\y) {};
    \node[] (L1) at (\layersep,-\nodesep*15-1*\nodesep) {$L^{(1)}$};

    \foreach \name / \y in {1,...,15}
        \path[yshift=0.5cm]
            node[hidden neuron softplus] (H2-\name) at (2*\layersep,-\nodesep*\y) {};
    \node[] (L2) at (2*\layersep,-\nodesep*15-1*\nodesep) {$L^{(2)}$};
            
    \foreach \name / \y in {1,...,15}
        \path[yshift=0.5cm]
            node[hidden neuron softplus] (H3-\name) at (2.5*\layersep,-\nodesep*\y) {};
    \node[] (Ll) at (2.5*\layersep,-\nodesep*15-1*\nodesep) {$L^{(l)}$};

    \node[output neuron, pin={[pin edge={->}]right:$u_1$}, right of=H3-3] (O-1) at (2.3*\layersep,-0.7)  {};
    \node[output neuron, pin={[pin edge={->}]right:$u_2$}, right of=H3-3] (O-2) at (2.3*\layersep,-1.5)  {};
    \node[below of=O-2] (Ll1) at (3.3*\layersep,-2*\nodesep) {$L^{(l+1)}$};

    \foreach \source in {1,...,5}
        \foreach \dest in {1,...,15}
            \path (I-\source) edge (H1-\dest);
            
    \foreach \source in {1,...,15}
        \foreach \dest in {1,...,15}
            \path (H1-\source) edge (H2-\dest);

    \path[thick] (H2-8) edge[-,loosely dotted] (H3-8);

    \foreach \source in {1,...,15}
        \foreach \dest in {1,...,2}
            \path (H3-\source) edge (O-\dest);

    \pgfmathsetmacro{\scale}{0.2}
    \pgfmathsetmacro{\Ox}{10}
    \pgfmathsetmacro{\Oy}{4}

    \draw[scale=\scale, shift = {(\Ox, \Oy)}] (-3,0) -- (3,0) node[right] {};
    \draw[scale=\scale, shift = {(\Ox, \Oy)}] (0,-1) -- (0,3) node[above] {softplus};
    \draw[scale=\scale,domain=-3:3,smooth,variable=\x,blue, shift = {(\Ox, \Oy)} ] plot ({\x},{ln(1+exp(\x)});
    
    \draw[scale=\scale, shift = {(\Ox+10, \Oy)}] (-3,0) -- (3,0) node[right] {};
    \draw[scale=\scale, shift = {(\Ox+10, \Oy)}] (0,-1) -- (0,3) node[above] {};
    \draw[scale=\scale,domain=-3:3,smooth,variable=\x,blue, shift = {(\Ox+10, \Oy)} ] plot ({\x},{ln(1+exp(\x)});
    
    \draw[scale=\scale, shift = {(\Ox+15, \Oy)}] (-1,0) -- (3,0) node[right] {};
    \draw[scale=\scale, shift = {(\Ox+15, \Oy)}] (0,-1) -- (0,3) node[above] {};
    \draw[scale=\scale,domain=-2:3,smooth,variable=\x,blue, shift = {(\Ox+15, \Oy)} ] plot ({\x},{ln(1+exp(\x)});
    
    \draw[scale=\scale, shift = {(\Ox+25, \Oy)}] (-1,0) -- (3,0) node[right] {};
    \draw[scale=\scale, shift = {(\Ox+25, \Oy)}] (0,-1) -- (0,3) node[above] {tanh};
    \draw[scale=\scale,domain=-2:3,smooth,variable=\x,red, shift = {(\Ox+25, \Oy)} ] plot ({\x},{tanh(\x)});
    
\end{tikzpicture}
\caption{The architecture of the G\&CNETs $\mathcal N$ studied. \label{fig:ffnn}}

\end{figure}
%
%
We indicate the optimal state feedback as approximated by a G\&CNET with $\mathcal N(\mathbf x) \approx \mathbf u^*(\mathbf x)$. Substituting this for $\mathbf u$ in the system dynamics gives:
\begin{equation}
\label{eq:nonlin}
\dot{\mathbf x} = \mathbf f(\mathbf x, \mathcal N(\mathbf x))
\end{equation}
$\mathcal N(\mathbf x)$ has a functional form that is hierarchical in its transformation from state to control. Intermediate computations can be considered on a layer-by-layer basis as we show below:

\begin{equation}
\label{eq:ann}
\mathcal N(\mathbf x): \enskip
\left\{
\begin{array}{l}
L^{(0)} = \mathbf g_{\textrm{pre}}(\mathbf x) \\
L^{(i+1)} = \sigma_i(\boldsymbol W^{(i)} L^{(i)} + \mathbf b^{(i)}), \enskip \forall i = 0..l \\
\mathcal N(\mathbf x) = \mathbf g_{\textrm{post}}(L^{(l+1)})
\end{array}
\right.
\end{equation}
%

The number of layers (network \textit{depth}) $l$ as well as the weight matrix $\boldsymbol W^{(i)}$ and bias vector $\mathbf b^{(i)}$ dimensionality across all layers (network \textit{width}) constitute the network architecture.
$\sigma_i$ is a non-linear function termed \textit{activation function} selected for each layer.
$\mathbf g_{\textrm{pre}}(\cdot)$ and $\mathbf g_{\textrm{post}}(\cdot)$ are transformation functions resulting from the data preprocessing step of the standard machine learning pipeline.
They are necessary for two reasons: (1) many machine learning algorithms have an assumption on the distribution of the training data; (2) to re-scale neural network outputs allowing for the use of arbitrary activation functions in the output layer.
It is worth stating that these functions are unchanged during learning and therefore the same transformation functions are used when we consider different network architectures in Sec.~\ref{sec:exp}.
The neural network parameters ($\{\boldsymbol W^{(i)}, \mathbf b^{(i)}\}_i$) are found during training, typically using some variant of stochastic gradient descent.
This stochasticity in the optimisation algorithm means that in repeated training runs the fitted parameters could be different and therefore properties of the neurocontrolled system (\ref{eq:nonlin}) could change. We note that even for networks with a single hidden layer, the expanded form shown in equation (\ref{eq:ann}) provides little mathematical insight on the resulting control structure.
Furthermore, its highly non-linear form gives rise to skepticism on the use of such controllers in applications where the satisfaction of requirements must be proved using formal mathematical tools. 
Despite this, it is worth stating that $\mathcal N(\mathbf x)$ is a
function in the differentiability class $\mathrm C^k$ where $k$ is the lowest amongst all the differentiability classes of its activation functions $\sigma_i$. This suggests that the straightforward deployment of techniques such as linearisation and even high-order Taylor map approximations is possible.

In this work, we restrict our attention to neural networks that use \emph{softplus} activation functions ($\sigma_{\textrm{soft}}(x) = \log(1+\exp(x))$) in the hidden layers and \emph{tanh} activation functions in the output layer (see Fig. \ref{fig:ffnn}). Consequently, the network outputs are continuous and differentiable everywhere since they belong to differentiability class $C^\infty$.
This avoids issues that could arise when the popular \emph{ReLU} activation function ($\sigma_{\textrm{relu}}(x) = \textrm{max}(0, x)$) is used.
We also note that for the dynamical model considered in section \ref{sec:exp}, comparable training performance was observed when softplus activation functions were used instead of ReLUs \cite{tailor2019learning}.
Since a G\&CNET is expected to approximate the optimal feedback $\mathbf u^*$, a good requirement on the neurocontrolled system described by equation (\ref{eq:nonlin}) is to also produce in $\mathbf x_e$ a globally stable equilibrium point.
%

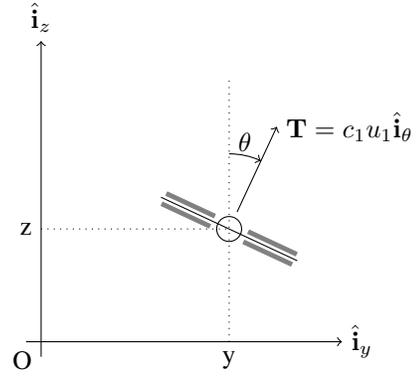
\begin{figure}[tb]
\centering
\begin{tikzpicture}[domain=0:4]
    \pgfmathsetmacro\th{-25}
    \pgfmathsetmacro\L{2}
    \pgfmathsetmacro\yltip{1}
    \pgfmathsetmacro\zltip{2}
    \pgfmathsetmacro\ycom{2.5}
    \pgfmathsetmacro\zcom{1.5}  
    \pgfmathsetmacro\eps{0.05}  

    \node[draw, circle] at (\ycom, \zcom)  (COM) {};
    \node[below left] at (0, 0)  (O) {O};
    \draw[->] (-0.2,0) -- (4,0) node[right] {$\hat{\mathbf i}_y$};
    \draw[->] (0,-0.2) -- (0,4) node[above] {$\hat{\mathbf i}_z$};
        \begin{scope}[rotate around={\th:(COM)}]
        \draw[-] (\ycom-\L/2,\zcom) -- (\ycom+\L/2,\zcom);
        \draw [fill=gray,gray] (\ycom-\L/2+\eps,\zcom+\eps) rectangle (\ycom-5*\eps,\zcom+2*\eps);
        \draw [fill=gray,gray] (\ycom-\L/2+\eps,\zcom-\eps) rectangle (\ycom-5*\eps,\zcom-2*\eps);
        \draw [fill=gray,gray] (\ycom+\L/2-\eps,\zcom+\eps) rectangle (\ycom+5*\eps,\zcom+2*\eps);
        \draw [fill=gray,gray] (\ycom+\L/2-\eps,\zcom-\eps) rectangle (\ycom+5*\eps,\zcom-2*\eps);
        \draw[->] (\ycom, \zcom+5*\eps) -- (\ycom, \zcom + 1.5) node[right] {$\mathbf T = c_1 u_1  \hat{\mathbf i}_\theta$};
    \end{scope}
    \draw[->] (\ycom, \zcom+1) arc (90:90+\th:1) node[above left] {$\theta$};
    \draw[dotted] (\ycom, 0) node[below] {y} -- (\ycom, \zcom + 2);
    \draw[dotted] (0, \zcom) node[left] {z} -- (\ycom, \zcom);
\end{tikzpicture}
\caption{The quadcopter system. \label{fig:quad}}
\end{figure}

\subsection{Neurocontroller dynamics linearisation}
\subsubsection{Linear stability of the equilibrium}

We stated earlier that $\mathbf x_e$ is a globally asymptotically stable equilibrium point for the optimally controlled system.
Here we demonstrate that the asymptotic stability of $\mathbf x_e$, at least in the local sense, for the system controlled by a G\&CNET can be verified.
We also emphasise that the representation of the optimal feedback by a neural network allows for the evaluation of linear stability margins.
This is not available for $\mathbf u^*(\mathbf x)$ as the extended definition has a discontinuity point at $\mathbf x_e$ where the control instantaneously takes the value $\mathbf u_e$.
This means the optimal dynamics in a neighbourhood of $\mathbf x_e$ does not follow laws of type $\exp(-ct)$ and thus cannot be described by a linear system.

Consider the neurocontrolled system in equation (\ref{eq:nonlin}) and derive, around its equilibrium point $\mathbf x_e$,  the linearized form $\dot {\mathbf x} = \mathbf A\mathbf x$ where $\mathbf A = \nabla \mathbf f\left|_{\mathbf x = \mathbf x_e}\right.$. 
Let us expand this expression further to highlight the relationship between the linearized dynamics and the neural network gradients. Indicating the matrix $\mathbf A$ through its components $a_{ij}$ we have:
%
%
\begin{equation}
\label{eq:linA}
a_{ij} = \frac{df_i}{dx_j} = \underbrace{\frac{\partial f_i}{\partial x_j}}_{a_{s,ij}} + \underbrace{\sum_{k=1}^{m} \frac{\partial f_i}{\partial u_k} \mathcircled{\frac{\partial \mathcal N_k}{\partial x_j}}}_{a_{\mathcal N,ij}}.
\end{equation}
This shows that the linearized dynamics is the sum of two matrices: $\mathbf A_s$ with components $a_{s,ij}$ representing the system dynamics, and $\mathbf A_{\mathcal N}$ with components $a_{\mathcal N,ij}$ representing the network feedback.
It is then clear that to compute the linear dynamics one needs to compute the derivatives of the neural network outputs with respect to its inputs. 
This task is most efficiently solved using automatic differentiation, a technique of widespread use in deep learning as it forms the core of the backpropagation algorithm used to train neural networks. 
As a consequence, there are an abundance of software frameworks that implement this (e.g. \texttt{Tensorflow}, \texttt{Pytorch}).

The dynamics of the neurocontrolled system in a local regime about the equilibrium point can be determined by the eigenvalues $\lambda_i  = \alpha_i + j \omega_i$ of $\mathbf A$. Briefly, an initial perturbation $\delta \mathbf x_0$ will evolve as $\delta \mathbf x = e^{\mathbf A t} \delta \mathbf x_0 $, where we have introduced the variable $\mathbf x = \mathbf x_e + \delta \mathbf x$. The solution $\delta \mathbf x = e^{\mathbf A t} \delta \mathbf x_0$ is a linear combination of terms in the form $e^{-\lambda_i t}$. Therefore, $\alpha_i$ defines the asymptotic behavior of the system and $\omega_i$ determines the frequency $f = \frac {\omega_i}{2\pi}$ of its oscillatory behaviour.

\subsubsection{Stability to time delay}
The principal reason for the use of G\&CNETs is for on-board, real-time optimal control \cite{sanchez2018real}.
In such contexts, the time scale to apply the optimal action is typically short.
This makes studying the neurocontroller behaviour with respect to feedback delays important for ensuring the robustness of the proposed controller.
To this effect, we introduce the time delay $\tau$ in the system dynamics as follows:
\begin{equation}
\label{eq:nonlintau}
\mathbf{\dot x(t)} = \mathbf f(\mathbf x(t), \mathcal N(\mathbf x^\tau))
\end{equation}
where $\mathbf x^\tau = \mathbf x(t-\tau)$. Essentially, we assume that the current action is calculated based on the system state delayed by $\tau$.
The linearized state space representation around $\mathbf x_e$ is:
\begin{equation}
\label{eq:lin_delayed}
\dot {\delta \mathbf x} = \mathbf A_s \delta \mathbf x + \mathbf A_{\mathcal N} \delta \mathbf x^\tau.
\end{equation}
This linear system with time delay has the general solution~\citep{Luo2017}: $\delta \mathbf x = e^{\mathbf \Phi t}\mathbf \delta \mathbf x_0$ where the eigenvalues of $\mathbf \Phi$ are obtained from the time-delayed characteristic equation:
\begin{equation}
\label{eq:timedel}
\left|\mathbf A_s + \mathbf A_N e^{-\lambda \tau} - \lambda\mathbf I\right| = 0.
\end{equation}
We note that for $\tau =0$ (no delay) we recover the stability study of the non-delayed system. The solution to the above equation forms the root locus of the system for different delay values. Detecting the first occurrence of the imaginary axis crossing by one of the roots allows us to determine the critical time delay, $\tau^*$, that destabilizes the system. 
In practice, however, the solution to equation (\ref{eq:timedel}) comes with quite some issues associated to the initial guess and the non-linear nature of the equation. While eigenvalue tracking approaches can help to alleviate such issues, the resulting numerical methods are cumbersome and inefficient.
As a consequence, in this work we use the Pad\'e approximation of order 5 for the time delay, to obtain the initial guess for solving equation (\ref{eq:timedel}).
The time delay in the Laplace domain can be approximated as a rational function of two polynomials of equal degree, $e^{-s \tau} \approx  P(s)/Q(s)$, where $s$ is the Laplace variable.
Transforming the approximation in the time domain and applying it separately on each state of the original system results in:
\begin{align}
\label{eq:ltiPade_out}
\delta \mathbf x^\tau & \approx \mathbf C_p \mathbf x_p + \mathbf D_p \delta \mathbf x \\
\label{eq:ltiPade_sys}
\dot{\mathbf x}_p & = \mathbf A_p \mathbf x_p + \mathbf B_p \delta \mathbf x
\end{align}
where the matrices $\mathbf A_p$, $\mathbf B_p$, $\mathbf C_p$ and $\mathbf D_p$ depend on $\tau$, and $\mathbf x_p$ are internal states of the system.
The input to this linear system is the current state $\delta \mathbf x$ and the output is the approximation of the delayed state $\delta \mathbf x^\tau$.
The interconnection of the delayed system (\ref{eq:lin_delayed}) with the approximation (\ref{eq:ltiPade_out}-\ref{eq:ltiPade_sys})
$$
\begin{bmatrix}
\delta \dot{\mathbf x} \\ \dot{\mathbf x}_p
\end{bmatrix}
=
\begin{bmatrix}
\mathbf A_s + \mathbf A_{\mathcal N} \mathbf D_p & \mathbf A_{\mathcal N} \mathbf C_p \\
\mathbf B_p & \mathbf A_p
\end{bmatrix}
\begin{bmatrix}
\delta \mathbf x \\ \mathbf x_p
\end{bmatrix}
$$
gives a linear system with an augmented state.
An estimate for $\tau^*$ can be obtained by calculating the eigenvalues of the system matrix.

\begin{figure}[tb]
\centering
\includegraphics[width=0.9\columnwidth]{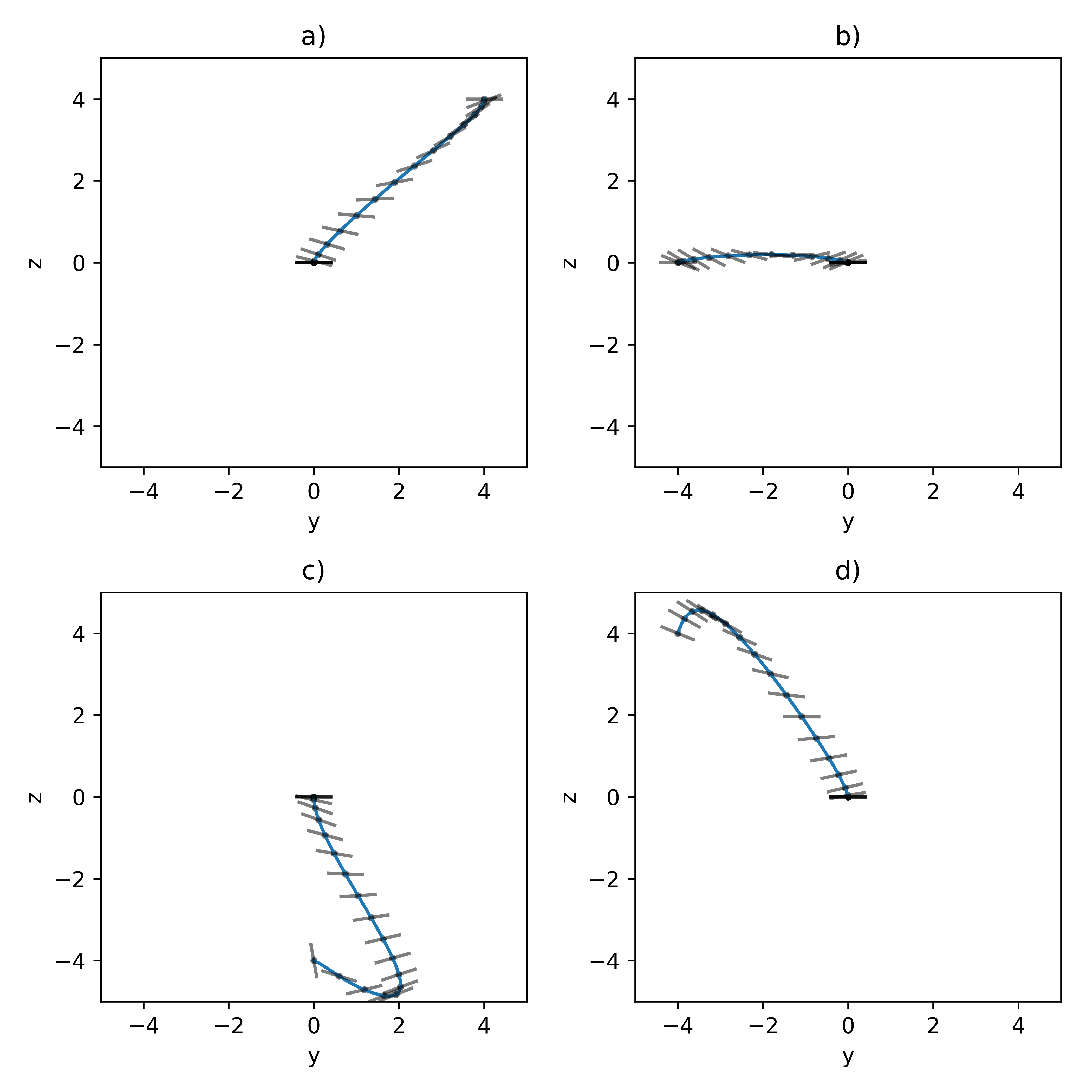}
\caption{Four different manoeuvres of the quadcopter controlled by a G\&CNET with the position and orientation variables indicated. The corresponding optimal trajectories are not visible due to overlap with the ones reported. \label{fig:trajs}}
\end{figure}

\subsection{High-order Taylor maps of neurocontrolled trajectories}
So far, we have only applied standard tools of control theory to the study of the system described by equation (\ref{eq:nonlin}). By doing so, we were able to look into the behaviour of a G\&CNET controlled system in a neighbourhood of the equilibrium point.
We now propose a method based on differential algebra and high-order Taylor maps \cite{berz1988differential} to study the solutions of (\ref{eq:nonlin}) in a neighbourhood of a reference solution (or nominal trajectory).
The basic idea is to use differential algebraic techniques to compute a high-order Taylor representation of such solutions. These expansions can then be be used to analyze the global behaviour of the neurocontrolled system to parametric uncertainties.
In contrast to the standard approach, we do not seek to investigate if small perturbations along a given tracked trajectory die out.
Instead, we look into the adjustments the neurocontroller makes as it is perturbed from an initial nominal trajectory defined by $\mathbf x_0$.
We note that this same technique can also be applied to the analysis of perturbations with respect to other model parameters.

Let us indicate the solution to the initial value problem of (\ref{eq:nonlin}) with:
$$
\mathbf x(t) = \Phi_{\mathcal N}(t, \mathbf x_0)
$$
where the subscript reminds us that the system dynamics is controlled via a neural network $\mathcal N$.
We may then compute its Taylor representation of order $k$ and write:
$$
\mathbf x(t)  \approx \mathcal M^{k,t}_{\mathbf x_0}(\delta \mathbf x_0).
$$
The map $\mathcal M^{k,T_j}_{\mathbf x_0}$ describes, to order $k$ and at time $t$, the solution to the initial value problem with initial condition $\mathbf x_0 + \delta \mathbf x_0$. 
Such representations are called Taylor models, or high-order Taylor maps, and are computed efficiently using differential algebraic techniques such as those implemented in the open source python library \texttt{pyaudi} \citep{audi} and used here. 
We compute such maps by integrating numerically, with a variable step Runge-Kutta-Fehlberg scheme, the dynamics of the system from the initial condition $\mathbf x_0 + \delta \mathbf x_0$ up to the time $T$. 
As the numerical integration is defined over generalized dual numbers \citep{audi} rather than floating point numbers, the resulting $\mathcal M^{k,T_j}_{\mathbf x_0}$ are computed at times $T_j$ defined by the adaptive step size procedure of the numerical integrator (see \citep{armellin2010asteroid} for a recent overview of differential algebra and its use to compute high-order Taylor maps, or \citep{berz1998verified} for an earlier work).

Each map $\mathcal M^{k,T_j}_{\mathbf x_0}$ is a $k$-th order Taylor polynomial in $\delta \mathbf x_0$ and can be used to study the neurocontroller behaviour around a nominal trajectory since it represents an explicit form of the solutions of (\ref{eq:nonlin}), at least in a ball $B_\epsilon$ of radius $\epsilon$ around the nominal trajectory initial conditions $\mathbf x_0$.
Consequently, starting from an initial perturbation in $B_\epsilon$, the asymptotic behaviour of the system will then be described by $\mathcal M^{\infty,t}_{\mathbf x_0}$ when $t \rightarrow \infty$.
This fact can be used to verify numerically if the neurocontrolled system will eventually reach a target point under the effect of external disturbances.
In the following we discuss the problem of providing an estimate for $\epsilon$.

\begin{table}[bt]
\centering
\caption{Bounds for random sampling of initial conditions}
\begin{tabular}{c@{\qquad}l}
  \toprule
  State variable & Interval \\
  \midrule
  $y$  & $\interval{-10}{10} \, \si{\metre}$ \\
  $z$  & $\interval{-10}{10} \, \si{\metre}$ \\
  $v_y$ & $\interval{-5}{5} \, \si{\metre\per\second}$ \\
  $v_z$ & $\interval{-5}{5} \, \si{\metre\per\second}$ \\
  $\theta$ & $\interval{-\frac{\pi}{4}}{\frac{\pi}{4}} \, \si{\radian}$ \\
  \bottomrule
\end{tabular}
\label{tab:IC4quad}
\end{table}

\subsubsection{Estimating the convergence radius of the Taylor model}
Let us introduce $\delta \mathbf x = \mathcal M^{k,t}_{\mathbf x_0} - \Phi_{\mathcal N}(t, \mathbf x_0)$, the evolution of the initial perturbation $\delta \mathbf x_0$.
In order to obtain an estimate for the convergence radius of this Taylor series, we bound the magnitude of $\delta \mathbf x$ as:
\begin{equation}\label{eq:deltax_norm}
\lVert \delta \mathbf x \rVert \le \lVert \mathbf A_1 \rVert \lVert \delta \mathbf x_0 \rVert + \lVert \mathbf A_2 \rVert \lVert\delta \mathbf x_0 \rVert^2 + \lVert \mathbf A_3 \rVert \lVert \delta \mathbf x_0 \rVert^3 + \ldots
\end{equation}
where $\mathbf A_i$ are the matrices resulting from unfolding the tensor of the $i$-th Taylor term, i.e. $i=1$ represents the gradient, $i=2$ the Hessian and so forth.
A detailed derivation of (\ref{eq:deltax_norm}) is given in the Appendix.
As a result for an expansion of order $k$, we can write:
$$
\lVert  \delta \mathbf x \rVert  \le \sum_{i=0}^k \lVert \mathbf A_i \rVert \epsilon^i = \sum_{i=0}^k b_i \epsilon^i
$$
where we introduced, for shortness, $b_i$ as the corresponding matrix norms and $\lVert \delta \mathbf x_0 \rVert = \epsilon$. We may then study the convergence radius of the series $\sum_{i=0}^\infty b_i\epsilon^i$ to obtain a conservative estimate of the radius of convergence of $\mathcal M^{\infty,t}_{\mathbf x_0}$. Applying the ratio test (D'Alembert's criterion) we have: $\epsilon \lim_{k \rightarrow \infty} \frac{|b_{k+1}|}{|b_k|} < 1$ and hence we can define:
\begin{equation}
\label{eq:lim}
\epsilon = \lim_{k \rightarrow \infty} \frac{|b_{k}|}{|b_{k+1}|}
\end{equation}
and guarantee that the series $\mathcal M^{\infty,t}_{\mathbf x_0}$ converges in the ball $B_\epsilon$.
Therefore, starting from initial conditions perturbed in $B_\epsilon$, $\mathbf x_e$ will be an asymptotically stable point in $B_\epsilon $ if and only if $\lVert \mathbf A_j \rVert \rightarrow 0 \quad \forall j$, as $t \rightarrow \infty$. Such a condition can easily be verified numerically up to $j=k$ once $\mathcal M^{k,t}_{\mathbf x_0}$ is computed for a sufficiently large $t$.

To summarize, the use of the techniques described in this section allow us to study with respect to G\&CNETs: a) the local stability of the equilibrium point $\mathbf x_e$; b) the linear time delay margin of the system; c) the stability of a nominal neurocontrolled trajectory originating from $\mathbf x_0$. In the next section we will use these techniques to study a practical test case: the two-dimensional dynamics of a quadcopter.

\section{Numerical experiments}
\label{sec:exp}

\subsection{The quadcopter model}
Consider the system shown in Fig.~\ref{fig:quad} representing a quadcopter whose two-dimensional dynamics is defined by the following set of ordinary differential equations:
\begin{equation}
\left\{
\begin{array}{l}
    \dot{\mathbf r} = \mathbf v \\
    \dot{\mathbf v} = c_1 \frac {u_1}{m} \hat{\mathbf i}_\theta +  \mathbf g - \frac 12 \mathbf v\\
    \dot{\theta} = c_2 u_2
\end{array}.
\right.
\label{eq:Q_ODE}
\end{equation}
The quadcopter state includes its position $\mathbf r = (y,z)$, its velocity $\mathbf v = (v_y, v_z)$ and its orientation $\theta$. We also refer to the state using the variable $\mathbf x = [\mathbf r, \mathbf v, \theta]$ and to the controls using the variable $\mathbf u = [u_1, u_2]$. In the rest of this paper, we use data from the \textregistered Parrot Bebop drone. The mass of the quadcopter is set to be $m = \SI{0.38905}{\kilo\gram}$ and the acceleration due to the Earth's gravity is $\mathbf g = (0, -g)$ where $g=\SI{9.81}{\meter\per\second\squared}$. Note that the dynamics of the quadcopter is also perturbed by a drag term and its coefficient is set to $\frac 12$. The control $u_1 \in [0,1]$ corresponds to a thrust action applied along the direction ${\mathbf i}_\theta = [\sin \theta, \cos \theta]$ bounded by a maximum magnitude $c_1 = \SI{9.1}{N}$. The control $u_2 \in [-1, 1]$ models a pitch rate bounded by $c_2 = \SI{35}{\radian\per\second}$. 

For the above non-linear dynamical system we consider the optimal control problem of steering the state from any initial state to the target state, $\{\mathbf r = (0,0), \mathbf v = (0,0), \theta = 0\}$ (i.e. $\mathbf x_f = \mathbf O$), minimizing the cost function: 
\begin{equation}
\begin{array}{l}
J = \int_0^{t_f} \left(c_1^2u_1^2 +c_2^2u_2^2\right) dt  \\
\end{array}
\end{equation}
We solve the resulting control problem from 200,000 initial conditions randomly sampled from the intervals reported in Table \ref{tab:IC4quad}.
For each optimal trajectory we insert in a database the state-action pairs at 59 equally spaced points, taking care to eliminate the last point and substitute it with 
$(\mathbf x_f, \mathbf u^*(\mathbf x_f))$ where $\mathbf u^*(\mathbf x_f) = [\frac{m g}{c_1}, 0]$ thereby ensuring the origin is an equilibrium point for the system (\ref{eq:Q_ODE}).
We then train neural networks with varying architectures $\mathcal N_l^\eta $, where $l$ is the network depth and $\eta $ is the network width held constant for all hidden layers, as a regression task on the database containing 11,800,000 optimal state-action pairs. 

We report in Table \ref{tab:mse} the performance (mean absolute error evaluated on a held-out partition of the training database) of networks with depth $l \in \{1..9\}$ and width $\eta \in \{50,100,200\}$.
For further details on the training procedure see \citep{tailor2019learning}.

\begin{table}[bt]
\caption{\label{tab:mse} Mean Absolute Error (x1000) and number of parameters for the various trained G\&CNETS $\mathcal N_l^\eta $}
\centering
$l$
\begin{tabular}{l r@{ / }r r@{ / }r r@{ / }r}
\multicolumn{7}{c}{$\eta $} \\
\hline
& \multicolumn{2}{c}{50} & \multicolumn{2}{c}{100} & \multicolumn{2}{c}{200} \\
\hline
1 & 31.7 & 402 & 30.3 & 802 & 28.2 & 1602\\
2 & 11.3 & 2952 & 8.4 & 10902 & 7.3 & 41802\\
3 & 7.9 & 5502 & 7.2 & 21002 & 6.6 & 82002\\
4 & 7.5 & 8052 & 6.1 & 31102 & 5.9 & 122202\\
5 & 7.9 & 10602 & 6.1 & 41202 & 5.9 & 162402\\
6 & 6.5 & 13152 & 6.4 & 51302 & 5.9 & 202602\\
7 & 6.3 & 15702 & 5.9 & 61402 & 5.9 & 242802\\
8 & 6.6 & 18252 & 6.0 & 71502 & 6.2 & 283002\\
9 & 7.4 & 20802 & 6.0 & 81602 & 6.0 & 323202\\
\hline
\end{tabular}
\end{table}

\begin{table}[t]
\caption{\label{tab:linres} Stability margins for various G\&CNETs}
\centering
\begin{tabular}{lccc}
& $\zeta_{10\%}$ & $T$ & $\tau^*$ \\
\hline
$\mathcal N^{50}_{1}$ & 2.27s & 2.41s & 0.137s \\
$\mathcal N^{50}_{2}$ & 1.66s & -- & 0.034s \\
$\mathcal N^{50}_{3}$ & 1.73s & 2.63s & 0.023s \\
$\mathcal N^{50}_{4}$ & 1.26s & 2.52s & 0.026s \\
$\mathcal N^{50}_{5}$ & 1.19s & 3.32s & 0.037s \\
$\mathcal N^{50}_{6}$ & 0.89s & 2.07s & 0.039s \\
$\mathcal N^{50}_{7}$ & 0.59s & 2.00s & 0.056s \\
$\mathcal N^{50}_{8}$ & 0.71s & 8.34s & 0.069s \\
$\mathcal N^{50}_{9}$ & 0.78s & 2.68s & 0.091s \\
\hline
\end{tabular}
\hskip.3cm
\begin{tabular}{lccc}
& $\zeta_{10\%}$ & $T$ & $\tau^*$ \\
\hline
$\mathcal N^{100}_{1}$ & 1.97s & 2.48s & 0.140s \\
$\mathcal N^{100}_{2}$ & 2.09s & 3.80s & 0.029s \\
$\mathcal N^{100}_{3}$ & 1.80s & 3.59s & 0.022s \\
$\mathcal N^{100}_{4}$ & 2.54s & -- & 0.032s \\
$\mathcal N^{100}_{5}$ & 1.52s & -- & 0.037s \\
$\mathcal N^{100}_{6}$ & 1.40s & 1.19s & 0.091s \\
$\mathcal N^{100}_{7}$ & 1.70s & 1.06s & 0.071s \\
$\mathcal N^{100}_{8}$ & 1.25s & 2.18s & 0.059s \\
$\mathcal N^{100}_{9}$ & 2.13s & 0.56s & 0.047s \\
\hline
\end{tabular}
\hskip.3cm
\begin{tabular}{lccc}
& $\zeta_{10\%}$ & $T$ & $\tau^*$ \\
\hline
$\mathcal N^{200}_{1}$ & 1.64s & 2.56s & 0.134s \\
$\mathcal N^{200}_{2}$ & 1.94s & 3.17s & 0.026s \\
$\mathcal N^{200}_{3}$ & 1.39s & 3.21s & 0.024s \\
$\mathcal N^{200}_{4}$ & 2.33s & 3.90s & 0.050s \\
$\mathcal N^{200}_{5}$ & 1.18s & 4.37s & 0.060s \\
$\mathcal N^{200}_{6}$ & 1.87s & 5.59s & 0.043s \\
$\mathcal N^{200}_{7}$ & 1.30s & 1.23s & 0.050s \\
$\mathcal N^{200}_{8}$ & 1.39s & 1.30s & 0.042s \\
$\mathcal N^{200}_{9}$ & 0.89s & 0.54s & 0.024s \\
\hline
\end{tabular}
\end{table}

A consequence of approximating the optimal feedback using neural networks is that the equilibrium point of the system (\ref{eq:Q_ODE}) as controlled by a G\&CNET $\mathbf u = \mathcal N_l^\eta (\mathbf x)$ may not be exactly at the origin.
In our experiments we find that the offset of the equilibrium from the origin is rather small, between $10^{-3}$ and $10^{-2}$, particularly when compared to the intervals used for sampling of the initial conditions.
In our case, we can make the origin the equilibrium point for all G\&CNETs simply by evaluating the true equilibrium point $\hat{\mathbf x}_l^\eta $ followed by shifting the axis i.e. $\mathbf x \rightarrow \mathbf x + \hat{\mathbf x}_l^\eta $.

We show in Fig. \ref{fig:trajs} a few trajectories resulting from the use of one particular G\&CNET $\mathcal N_3^{100}$ to control the quadcopter dynamics.
The optimal trajectories, as evaluated by solving the two-point boundary value problem resulting from the application of Pontryagin's maximum principle, are indistinguishable from the ones produced by the network.

\subsubsection{Hovering stability and stability to time delay}
After shifting our axis, the quadcopter dynamics (\ref{eq:Q_ODE}) controlled by each G\&CNET $\mathbf u = \mathcal N_l^\eta(\mathbf x)$ has in the origin $\mathbf O$ its equilibrium point and thus $\mathcal N_l^\eta (\mathbf O) = [\frac{m g}{c_1}, 0]$. 
This corresponds to the quadcopter hovering over the origin and cancelling exactly the gravitational pull with its upward thrust. We thus proceed to study the behavior of the quadcopter close to the hovering conditions by computing all the eigenvalues $\lambda_i$ of $\mathbf A$ resulting from (\ref{eq:linA}), for each of the trained networks. In all cases, the modes computed resulted to be stable, that is $\alpha_i < 0 \; \forall i=1..5$, proving that all trained networks result in a stable dynamics near the hovering position. 
For each eigenvalue, we then proceeded to compute the decay time, $\zeta_{{10\%}_i} = \frac{\ln(0.1)}{\alpha_i}$, which is defined as the time the envelope associated to the $i$-th mode takes to decay to 10\% of its starting value, and the oscillating period, $T_i = \frac{2\pi}{\omega_i}$, associated to the $i$-th mode. In Table \ref{tab:linres} we report the largest values for $\zeta_{{10\%}}$ and $T$ for each network $\mathcal N_l^\eta $. At most one pair of complex conjugate eigenvalues was present in all cases tested. In three cases no oscillatory behaviour is present as all eigenvalues are real.

\begin{figure}[tb]
\centering
\includegraphics[width=0.62\columnwidth]{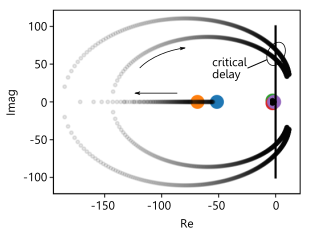}\\[1em]
\includegraphics[width=0.62\columnwidth]{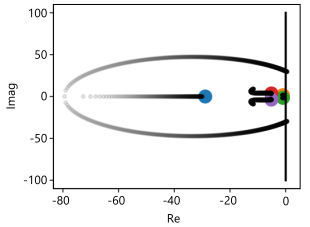}
\caption{Root locus for $\mathcal N_{100}^3$ (top) and $\mathcal N_{200}^4$ (bottom). The arrows indicate the movement of the eigenvalues for increasing $\tau$. \label{fig:rootlocus}}
\end{figure}

We then perform the analysis of the time-delayed system solving (\ref{eq:timedel}) for different values of the time delay $\tau$ and thus producing a root locus which we show in Fig.~\ref{fig:rootlocus} for two G\&CNETs: $\mathcal N_{100}^3$ and $\mathcal N_{200}^4$. 
The eigenvalues of the non-delayed system are also indicated in the figure. 
This shows that time delays initially have a stabilizing effect on the most stable roots but later give raise to instability represented by the crossing of the imaginary axis.
The computed values for the critical time delay $\tau^*$ are reported in Table \ref{tab:linres}.

From previous work \cite{sanchez2018real, tailor2019learning}, we know that neural networks with greater depths and widths (i.e. increased parameterisation) are better representations of the optimal feedback. 
This is repeated in Table \ref{tab:mse} for which we observe the largest reduction in error when the network goes from 1 to 2 hidden layers.
However, the stability margins characterising the neurocontroller behaviour in a neighbourhood of the equilibrium point (columns 1 and 2 of Table \ref{tab:linres}) indicate no clear relationship to the network architecture (and thus by extension the optimal feedback approximation performance).
One could argue that this is not surprising for two reasons, firstly the metric in Table \ref{tab:mse} is a measure of global performance, and secondly, as noted in section \ref{sec:methods}, the neurocontroller behaviour in a neighbourhood of the equilibrium point is not expected to match the optimal behaviour.
Despite this, we observe the critical time delay (column 3 of Table \ref{tab:linres}) for G\&CNETs with depth 1 to be considerably higher than the others.
In the context of the on-board implementation of G\&CNETs, this presents an interesting choice between greater optimality or improved robustness to time delays.

\subsubsection{High-order Taylor maps for the neurocontrolled quadcopter.}
\begin{figure}[tb]
\centering
\includegraphics[width=0.9\columnwidth ]{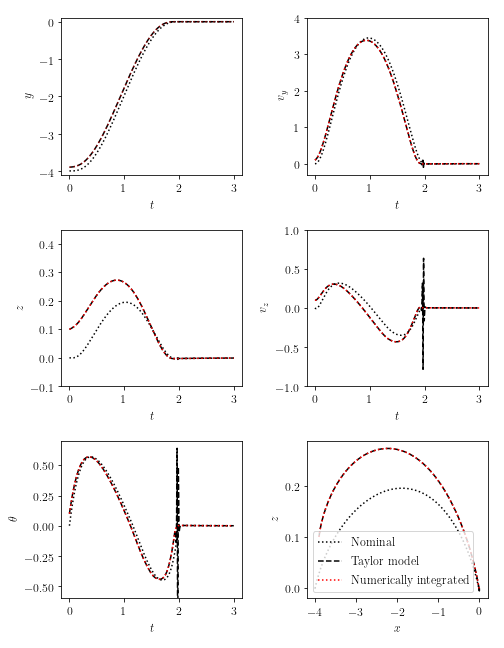}
\caption{Taylor model representation of an optimal trajectory \label{fig:taylormodel}. $\mathcal M^{7,i}_{\mathbf x_0}$ is shown for $\delta \mathbf x_0 = [0.1,0.1,0.1,0.1,0.1]\,\si{\meter}$.}
\end{figure}
Let us consider, as an example, the nominal trajectory of the quadcopter resulting from the neurocontroller $\mathcal N_{100}^3$ and the initial condition $\mathbf x_0 = [-4, 0,0,0,0]\,\si{\meter}$. The trajectory, also shown in Fig.~\ref{fig:trajs}b, is a manoeuvre where the quadcopter changes its position horizontally by \SI{4}{\meter}. We proceed to study the stability of such a nominal trajectory with respect to perturbations of the initial conditions $\mathbf x_0 + \delta \mathbf x_0$. In particular we would like to have an indication on the neurocontroller being able to bring the quadcopter to the final target state for all perturbations in some ball $B_\epsilon$. Note that, for this manouvre, the corresponding optimal time is $t^*_f = \SI{2.06}{\second}$.
Following the approach detailed in section \ref{sec:methods}, we use differential algebraic techniques and a Runge–Kutta–Fehlberg adaptive step numerical integration scheme to compute the high-order Taylor maps $\mathbf x_T \approx  \mathcal M^{k,T_j}_{\mathbf x_0}(\delta \mathbf x_0)$ representing the Taylor expansion of order $k$ of the quadcopter state at the time grid points $T_j$ defined by the adaptive stepper of the numerical integration scheme. The Taylor expansion is taken with respect to perturbations $\delta \mathbf x$ of the nominal initial conditions $\mathbf x_0$. 
Since $n=5$, i.e. the state has a dimension of five, each resulting Taylor polynomial contains 5, 20, 55, 125, 251, 461, 791 terms as the order grows from linear. 
Clearly, if: $\lim_{k\rightarrow \infty, t\rightarrow \infty} \mathcal M^{k,t}_{\mathbf x_0}(\delta\mathbf x_0) = \mathbf x_e$ then the neurocontroller is guaranteed to be able to drive the quadcopter to the final desired hovering position  $\forall \delta \mathbf x_0 \in B_\epsilon$.
When this is the case, a formal guarantee on the stability of the nominal trajectory is also obtained. 
Taking a numercial approach, since we cannot compute the Taylor model for $t \rightarrow \infty$, nor can we compute the map at order $k \rightarrow \infty$, we stop the numerical integration at $T = \frac 32 t_f^*$ (with $t^*_f$ representing the time of the optimal manoeuvre) giving the system enough time to reach the equilibrium position and stabilize itself. Inspecting the Taylor map computed for $t=T$ reveals that $\lVert \mathbf A_j \rVert < 1e^{-3} \quad \forall j \in [1 .. 7]$, and allows us to conclude that the equilibrium position is indeed acquired, with the said precision, from all initial conditions in an $\epsilon$ ball around the nominal $\mathbf x_0$. The ball size $\epsilon$ is then computed from (\ref{eq:lim}) using $k=7$.
\begin{figure}[tb]
\centering
\includegraphics[width=0.9\columnwidth]{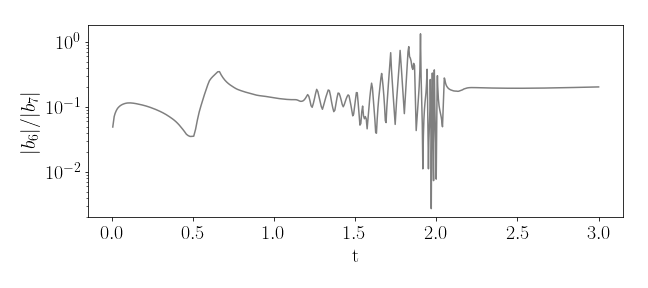}
\caption{Proxy for the convergence radius $\epsilon$ computed for each $\mathcal M^{7,T_j}_{\mathbf x_0}$. \label{fig:conv}}
\end{figure}
We can actually estimate the convergence radii for the map at all $T_j$. The result is shown in Fig.~\ref{fig:conv} where $\epsilon \approx \frac{|b_6|}{|b_7|}$ is plotted for the map $\mathcal M_{\mathbf x_0}^{7,t}$ against $t$ (the times $t$ are sampled in the grid $T_j$ defined by the adaptive stepper of the numerical integration scheme). The convergence radius at $T = \frac 32 t_f^*$ is approximately 0.1.
Around the optimal time $ t^*_f$ we also observe a dramatic decrease of the estimated convergence radius of the map. 
In order to confirm this observation we simulate in Fig.~\ref{fig:taylormodel} the quadcopter state during the nominal trajectory and the state predicted by the various maps $\mathcal M^{7,T_j}_{\mathbf x_0}(\delta \mathbf x_0)$ for a rather large perturbation $\delta \mathbf x_0 = [0.1,0.1,0.1,0.1,0.1]\,\si{\meter}$.
We also simulate the ground truth by numerical integration of (\ref{eq:nonlin}) using the initial conditions $\mathbf x_0 + \delta \mathbf x_0$. 
As expected, the various maps represent the new optimal trajectory, corresponding to the perturbed initial conditions, quite accurately for most time instants, failing catastrophically for $t \approx \SI{2}{\second}$, where the selected perturbation is evidently larger than the corresponding map convergence radius.

\section{Conclusions}
We show a general methodology to analyse the behaviour of neurocontrollers. Our method enables the study of neurocontrolled feedback dynamics in proximity of the equilibrium point by deriving linear stability margins as well as time delay margins. We then propose the use of high-order Taylor models to study the robustness of a nominal trajectory to perturbations on its initial conditions. The methodology is successfully applied to the test case of a two-dimensional quadcopter dynamical model where we show, for several G\&CNETs with varying architectures, the formal stability guarantees for the neurocontrolled hovering behaviour as well as for the stability of a nominal manoeuvre. We found that as soon as the network is deep, i.e. has more than one hidden layer, there seem to be no relation between the linear stability margins obtained and the network architecture. We also found that the computed Taylor model of the state describes well the neurocontrolled trajectories in a neighbourhood of a nominal trajectory and we propose a method to estimate the size of said neighbourhood. 
Our results constitute a first step to increasing the trust on the on-board use of an optimal feedback controller represented by a neural network, thus narrowing the gap between control theory and machine learning.

\appendix[Tensor unfolding]\label{app:tensorunfold}
Let us write the $i$-th component of $\delta \mathbf x = \mathcal M^{k,t}_{\mathbf x_0} - \Phi_{\mathcal N}(t, \mathbf x_0)$, namely the time evolution of an initial perturbation $\delta \mathbf x_0$, using the Einstein notation
\begin{multline*}
\delta x_i = a_{1,ij} \delta x_{0,j} + a_{2,ijk} \delta x_{0,j} \delta x_{0,k} + \\ a_{3,ijkl} \delta x_{0,j} \delta x_{0,k} \delta x_{0,l}  + \dots 
\end{multline*}
where $\delta x_{0,i}$ denotes the $i$-th component of $\delta \mathbf x_0$. The terms $a_1$, $a_2$ and $a_3$ are components of the gradient, the Hessian and the third-order partial derivatives tensors. 
We take the norm of the previous relation and we apply the triangle inequality and the properties of the induced matrix norms: 
\begin{equation*}
\lVert \delta \mathbf x \rVert \le \lVert \mathbf A_1 \rVert \lVert \delta \mathbf x_{0} \rVert + \lVert \mathbf A_2 \rVert \lVert\delta \mathbf x_{02} \rVert + \lVert \mathbf A_3 \rVert \lVert \delta \mathbf x_{03} \rVert + \ldots
\end{equation*}
To write the previous equation in a matrix-vector notation, we have unfolded the tensors such that all the partial derivatives corresponding to the same $\Phi_{\mathcal N, i}$ are in the same row, e.g. for $\mathbf A_{2,ij} = a_{2,ikl}$ and $\mathbf A_{3,ij} = a_{3,iklq}$ with $k=(j-1)\mod n+1$, $l=(j-1) \div n+1$ and $q=(j-1) \div n^2+1$, $j = 1, 2, \ldots, n^2$ and $j = 1, 2, \ldots, n^3$ for each matrix respectively.
We apply the same unfolding to the vectors $\delta \mathbf x_{02}$ and $\delta \mathbf x_{03}$ as  $\delta \mathbf x_{02,j} = \delta x_{0,k} \delta x_{0,l}$ and $\delta \mathbf x_{03,j} = \delta x_{0,k} \delta x_{0,l} \delta x_{0,q}$ where the indices $k$, $l$ and $q$ depend on $j$ as before.
Therefore, $\lVert \delta \mathbf x_{02} \rVert = \lVert \delta \mathbf x_0 \rVert^2$ and $\lVert \delta \mathbf x_{03} \rVert = \lVert \delta \mathbf x_0 \rVert^3$ under any $p$-norm.

\bibliographystyle{IEEEtran}
\bibliography{main}

\end{document}